\documentclass[letterpaper, 10 pt, conference]{ieeeconf}  

\IEEEoverridecommandlockouts                              

\usepackage{graphics} 
\usepackage{epsfig} 
\usepackage{mathptmx} 
\usepackage{times} 
\usepackage{amsmath} 
\usepackage{amssymb}  
\usepackage{graphicx}
\usepackage[table]{xcolor} 
\usepackage{cite,comment}
\usepackage{caption}    
\usepackage{subcaption}
\usepackage{multirow}
\usepackage{soul}
\usepackage{booktabs}

\definecolor{orange}{RGB}{252, 130, 62}
\definecolor{red}{RGB}{255, 0, 0}
\definecolor{brown}{RGB}{155, 25, 10}
\definecolor{blue}{RGB}{0, 0,255}
\definecolor{green}{RGB}{78, 196, 164}
\newcommand{\im}{\color{orange}}  
\newcommand{\mw}{\color{green}}  
  
\newcommand{\red}{\color{red}}

 \title{\LARGE \bf
 Textile Taxonomy and Classification Using Pulling and Twisting
 }

\author{Alberta Longhini, Michael C. Welle, Ioanna Mitsioni and Danica Kragic
\thanks{$^{1}$The authors are with the Robotics, Perception and Learning Lab, EECS, at KTH Royal Institute of Technology, Stockholm, Sweden
        {\tt\small albertal, mwelle, mitsioni, dani@kth.se}}%

}

\begin{document}

\maketitle
\thispagestyle{empty}
\pagestyle{empty}

\begin{abstract}
Identification of textile properties is an important milestone toward advanced robotic manipulation tasks that consider interaction with clothing items such as assisted dressing, laundry folding, automated sewing, textile recycling and reusing. Despite the abundance of work considering this class of deformable objects, many open problems remain. These relate to the choice and modelling of the sensory feedback as well as the control and planning of the interaction and manipulation strategies. Most importantly, there is no structured approach for studying and assessing different approaches that may bridge the gap between the robotics community and textile production industry.  To this end, we outline a textile taxonomy considering fiber types and production methods, commonly used in textile industry. We devise datasets according to the taxonomy, and study how robotic actions, such as pulling and twisting of the textile samples, can be used for the classification. We also provide important insights from the perspective of visualization and interpretability of the gathered data. 

\end{abstract}

\section{Introduction}

Interaction with deformable objects is an integral part of our everyday life but still a challenge for robotic systems. Work on robotic handling of textile or fabric traces back several decades \cite{1990} and, despite the clear need in industry and domestic applications, many of the problems related to perception, planning and control remain open. From the industrial perspective, textile production and subsequent processes of garment design in fashion industry are largely not automated. Fashion industry is also undergoing an important transformation to address sustainability concerns, given that textile and clothing overproduction has a significant negative impact on the environment.

\begin{figure}[t]
  \centering
  \includegraphics[width=.8\linewidth]{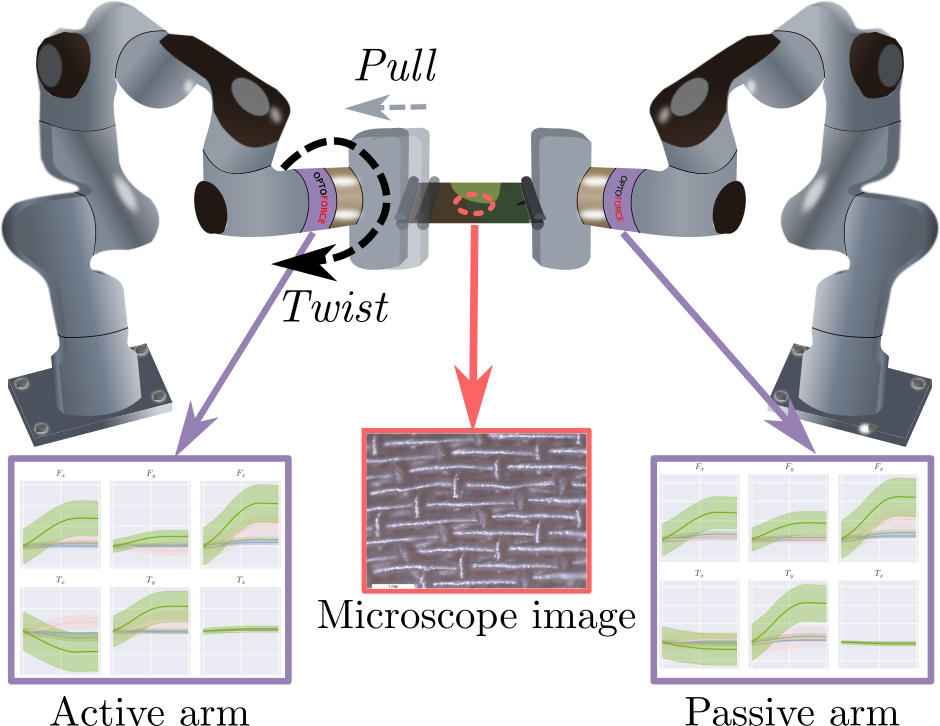}
  \caption{System setup: Two 7 DoF Franka Emika Panda arms with force-toque sensors on the wrists, twisting and pulling a textile sample. Microscope images are used to define textile classes prior to training.}
   \label{fig:system}
\end{figure}

From the scientific perspective, robotic interaction with deformable materials has gained significant attention recently \cite{deformableSurvey,doumanoglouFolding}. Important milestones regarding the modelling, perception, planning, control and simulation of deformable materials have been identified but not yet reached. It may even be so that until robots reach the dexterity, flexibility and sensing that to some extend resembles human capabilities, successful interaction with deformable objects will remain a challenge. In robotics, textile has been used to study manipulation tasks like folding \cite{LiFolding, doumanoglouFolding, lippi2020latent}, robot-assisted dressing \cite{dressing_dataDriven, dressingDynamics, garcia2020benchmarking},  garment recognition and classification  \cite{liu2016deepfashion, huang2015cross, ge2019deepfashion2, ziegler2020fashion}. In most of these works, only a subset of textile properties is commonly considered, and textile is merely a tool for testing sensors \cite{materialTactileSensing} or control strategies \cite{petrik2019feedback}.

In our work, we aim to study textile materials and their properties using physical interactions and wrist-mounted force-torque sensing. Similarly to humans, we aim at using actions such as pulling and twisting, to learn more about the textile properties, see Fig.~\ref{fig:system}. 
The properties are defined using a textile taxonomy that follows the classification used in the textile production industry. Textile properties in general, and thus interaction dynamics, are affected by factors such as fiber material and production method - the fiber may be raw, coated or it may be a blend of several materials.  Once used to produce garments or bed-clothing, the properties will change overtime based on washing, wearing, steaming - the textile can become harder or softer, less or more elastic, thinner. The change in properties will also have an impact on the planning and control strategies used to interact with it - how we wash, iron and fold them, how we hold and manipulate garments when dressing somebody, whether we decide to recycle or reuse old garments.  

The focus of this paper is to asses how a dual arm robotic system can be used to identify textile production methods through pulling and twisting.
We propose to do so by learning a classifier on a dataset of textile samples that are  annotated by their construction type, determined by inspecting their microscopic structure.
We first outline a textile classification taxonomy related to both fiber type and production method, following notation used in textile industry. We then make a thorough study using a subset of materials and production techniques to assess the validity of our approach. We analyse two manipulation strategies as well as investigate which measurements are most relevant for classification. We conclude by discussing challenges and open problems. 

\begin{figure*}[!t]
  \centering
  \includegraphics[width=\textwidth]{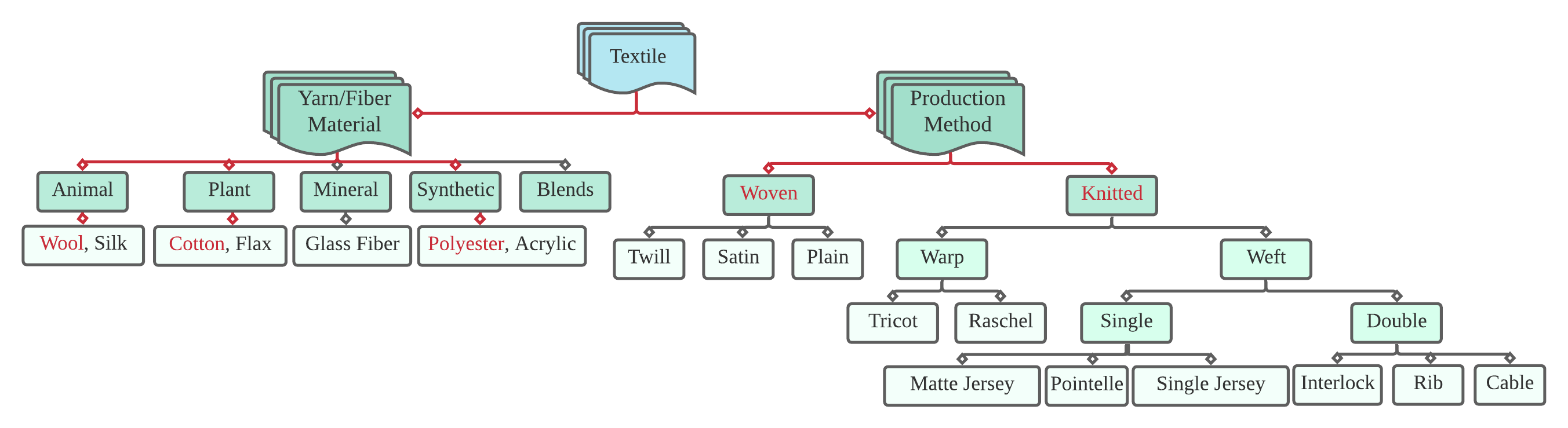}
  \caption{Textile taxonomy considering yarn/fabric material and production method. Classes considered in this work are  highlighted in red. 
  }\label{fig:taxo}
  \vspace{-0.2cm}
\end{figure*}

\section{Textile Taxonomy}
{\label{sec:taxonomy}}

Textile or fabric, is a deformable and flexible material 
made out of yarns or threads, which are put together by a construction or manufacturing process such as weaving, knitting, crocheting, knotting, tatting, felting, bonding or braiding. Most of the everyday clothing items we wear are constructed through weaving or knitting prior to sewing, although in high fashion other processes are used frequently too. Yarns and threads are produced by spinning raw fibers that may have different origin: animal, plant, mineral, synthetic or their blend. We summarize some of these aspects in Fig.~\ref{fig:taxo}, showing also some of the further distinctions in terms of differences in the manufacturing process.

Woven fabric is usually produced by using two sets of yarn, while knitted fabric employs a single set. To produce woven fabric, the yarn is interlaced, as opposed to knitting where it is interloped. Due to its construction, woven fabric is often hard and nonelastic, allowing it to hold creases well, and can be only stretched diagonally if the yarn itself is not a blend that includes elastic material. It is commonly used to produce garments such as shirts and jeans. 
On the other hand, knitted fabric is soft and can be stretched in all directions, making it rather wrinkle-resistant. One example of its frequent use is for t-shirts. One important aspect is that it usually does not stretch equally in all directions - for example, a t-shirt will stretch more horizontally than vertically to more naturally follow body shape. A more complete account of the properties of fabric can be found in \cite{GRISHANOV201128}.

The above is of importance for various robotics applications that consider active interaction with the textile. For assisted dressing applications, it is important for the robotic system to generate relevant control strategies when pulling up pants, helping with the sleeves or pulling down the t-shirt: hard textile may require completely different manipulation strategies and safety considerations than the flexible one. Most of the clothing items will have a content label attached to them and may offer information about the type of yarn used. However, the manufacturing process is never described on the label, neither is the fact that a clothing item may be a combination of woven and knitted parts and a combination of different type of textiles  being put together in the sewing step. For example, both a pair of jeans and a t-shirt may be made with 100\% cotton textile that, in the first case, is woven and in the second case, knitted. With the proposed taxonomy and the work in this paper, we report on some initial insights on how some of the textile properties can be examined by using force-torque measurements and actions such as pulling and twisting of the textile samples. 

\begin{figure}[!htb]
    \centering
  \includegraphics[width=.8\linewidth]{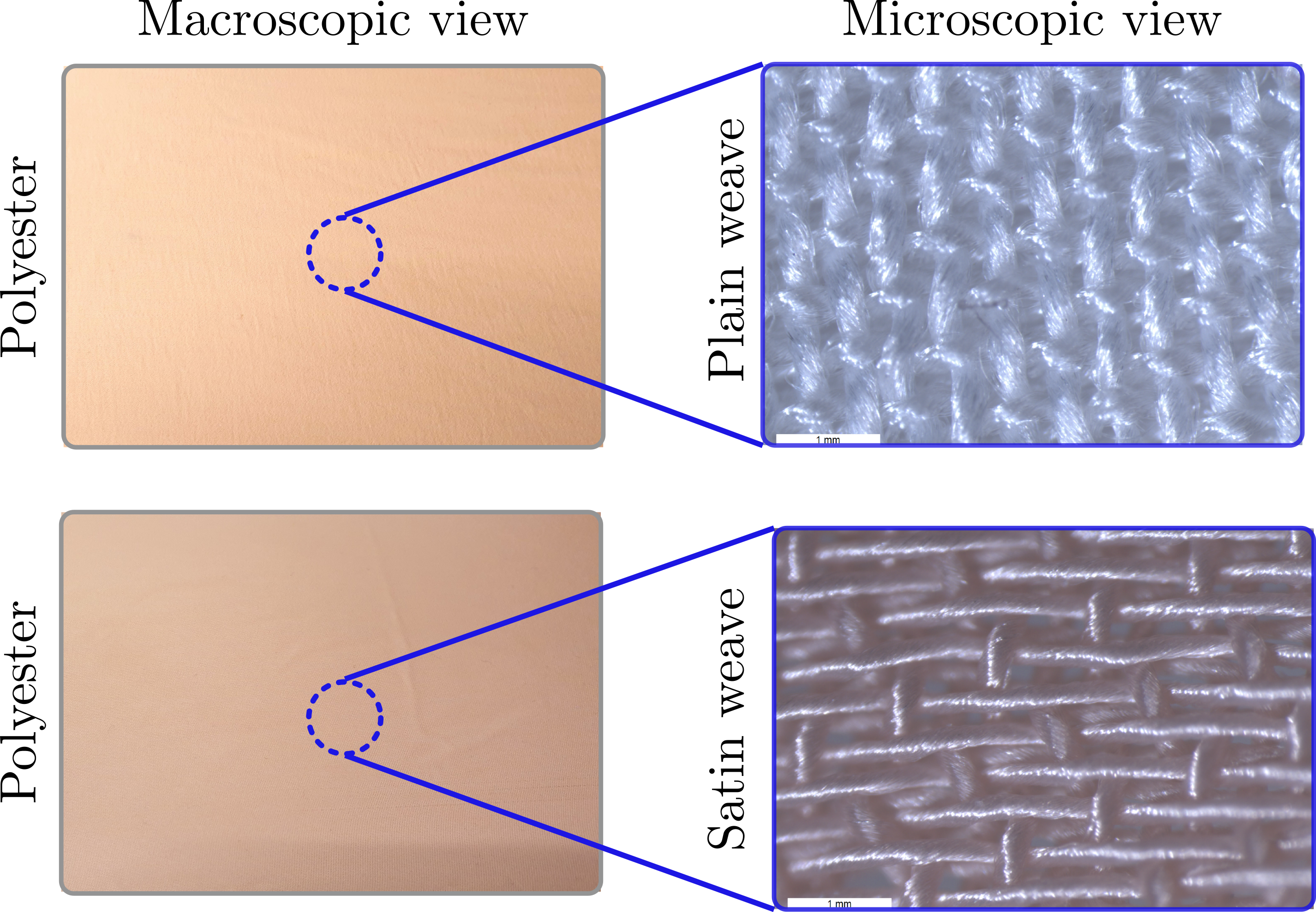}
    \caption{Textile can look the same given regular camera (left), but very different on the microscopic level (right). The different weaving styles (plain weave and satin weave) with the same material determine the dynamical properties that are important when manipulating fabric.}
    \label{fig:micro_fabric_different}
\end{figure}

\begin{figure}[!htb]
    \centering
  \includegraphics[width=.8\linewidth]{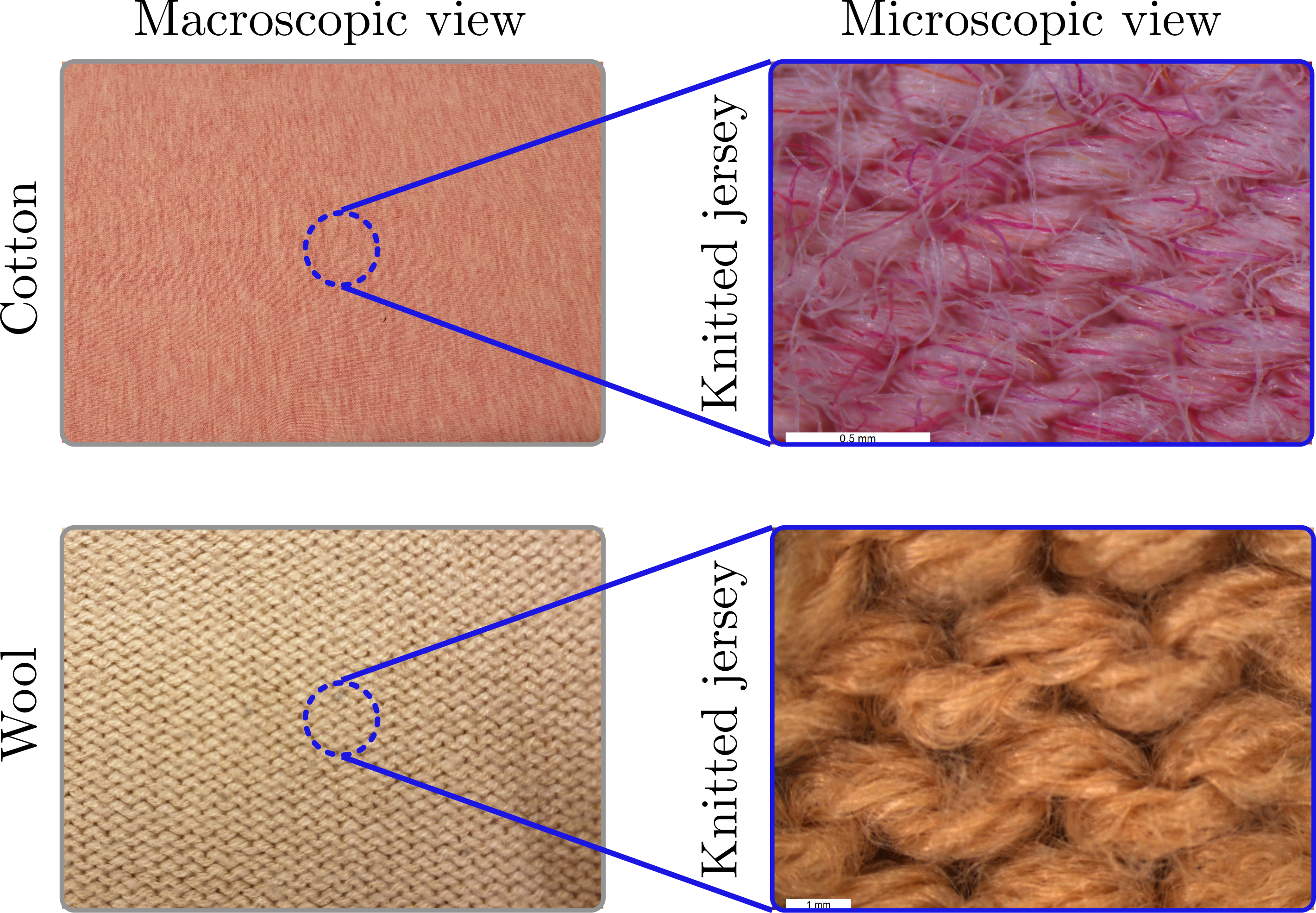}
    \caption{Textile can look different on the macro level, but very similar on the micro level. Left) two different materials (Wool and Cotton) using a regular camera. The microscope image (right) reveals that they have the same underlying construction - knitted jersey.}
    \label{fig:micro_fabric_equal}
    \vspace{-0.8mm}
\end{figure}

Even for humans, the manufacturing method may not be visible with the naked eye and the label only provides the yarn material. We use our experience of previously interacting with clothing items to choose appropriate actions when dressing ourselves or others, washing, ironing, repairing or sewing. To shed some light on this, we collected microscope images of our textile samples. While a regular camera image may not provide enough signal resolution in actual robot interaction with the textile, the high-resolution microscope images show different ways of interlacing yarn that has a huge effect on the elasticity of the textile Fig.~\ref{fig:micro_fabric_different}. Similarly, garments with different reflection or texture properties may look rather different under a regular camera but their dynamical properties may be the same if the manufacturing method, and yarn type, is the same, see Fig.~\ref{fig:micro_fabric_equal}.

In this paper, we therefore set out to investigate how force-torque measurements together with actions such as pulling and twisting may be used to classify the textiles according to the proposed taxonomy. We chose pulling and twisting since these are also the two most common actions humans use for inspecting textile properties.

\section{Related Work}
In robotics, textile materials have been considered from perception, learning, planning and control perspectives. Most notable applications consider folding, assisted dressing  or material classification \cite{ perceptionAssistiveManip, deformableSurvey}. 




Despite the broad interest in the computer vision community, most works concentrate on building clothing item taxonomies \cite{liu2016deepfashion} rather than identifying material properties. Problems such as garment motion prediction \cite{Patel_2020_CVPR}, classification \cite{huang2015cross},  dressing 3D simulated humans \cite{Bhatnagar_2019_ICCV}, have also been addressed. It has also been shown that wrinkle detection may be helpful for classification \cite{Sun_cloth_category, smoothnessVision}. However, with vision alone it may be difficult to estimate the physical attributes of textiles \cite{Luo2017RoboticTP} although the results in \cite{Davis_2015_CVPR} indicate that vibrations captured in video can be correlated to the stiffness and density of fabrics. 


In robotics, identifying textile properties is important, but there is no common taxonomy that allows for comparison and benchmarking of the proposed approaches. Recent work in \cite{Strese2020HapticMA} proposes a taxonomy of 184 materials including leather, fur and plant fiber but there is no focus on textile in particular or the production method. Haptic feedback has often been used to label various types of materials \cite{fishel2012bayesian, materialTactileSensing, haptic_surface_recog}. The authors in \cite{MultiChannelNN} study compliance and texture to classify 32 materials including textile. Non-contact techniques have been used in \cite{spectroscopy} to distinguish among five material categories, one of which was textile. Thus, none of these works focuses specifically on textile material, or considers fiber and production method in particular. When considering textile classification, it has been studied from the fiber material perspective \cite{fabricClassification, dressingDynamics} or properties such as thickness, softness and durability \cite{materialPerception}. Material texture identification has been addressed in \cite{vibritactile_humanoid,  Drigalski2017TextileIU,Luo2018ViTacFS}, without considering the difference between fiber material and textile production method. 
Given these, we believe that our initial study and outlined taxonomy provides examples of how textile classification can be studied in a more structured manner.

\section{Data collection and dataset design}
\begin{figure}[!htb]
  \centering
  \includegraphics[width=\linewidth]{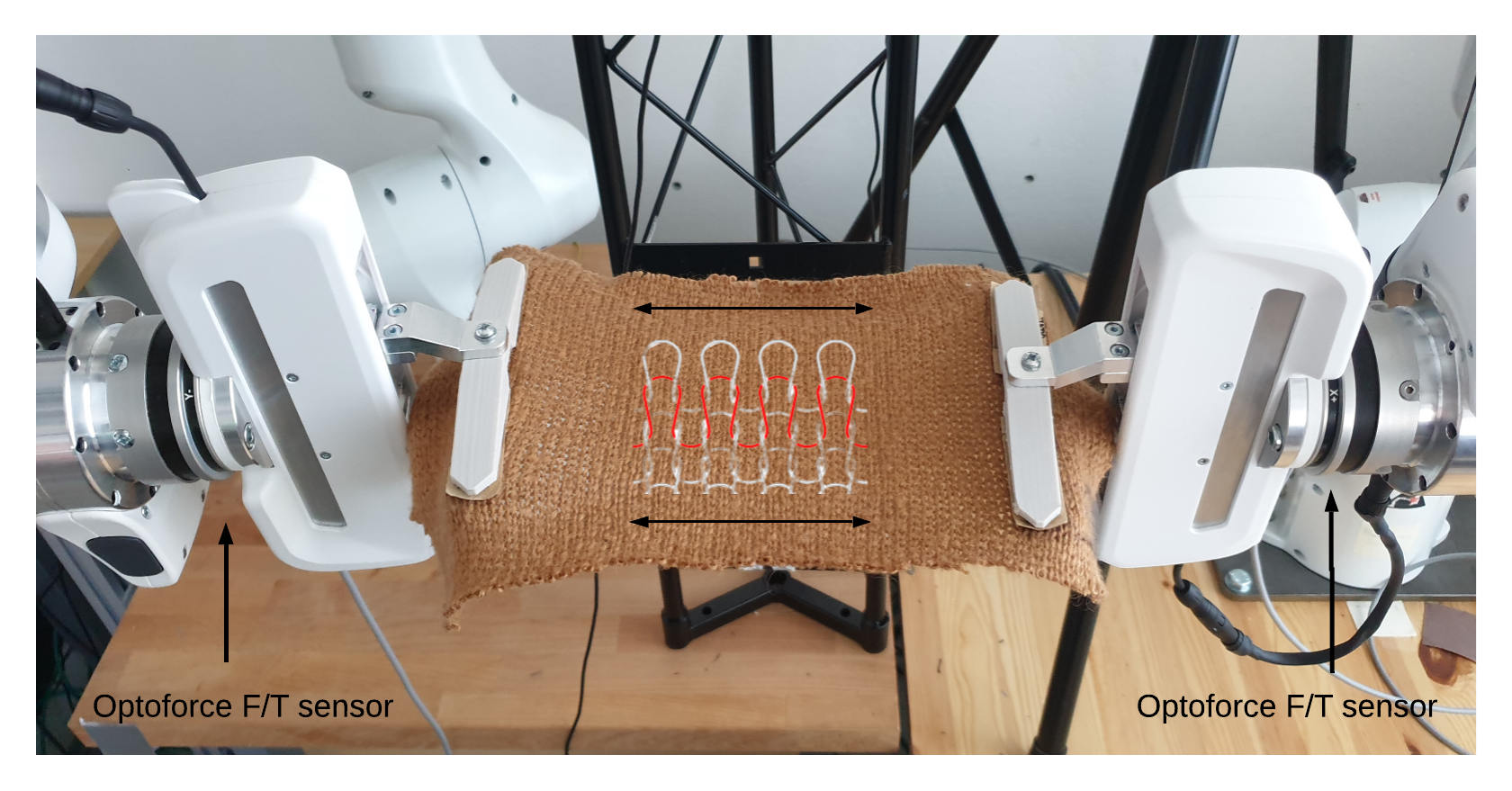}
   \includegraphics[width=\linewidth]{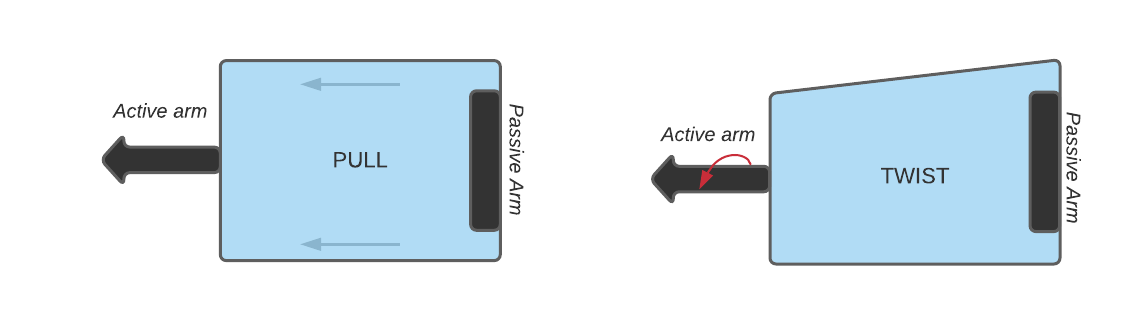}
  \caption{Upper) Flat 3D-printed grippers holding wool sample. The yarn direction coincides with the pulling direction. FT sensors are mounted on the wrist of the manipulators. Lower) Schematic example of \textit{pull} and \textit{twist}.}
   \label{fig:streching_explanation}
\end{figure}

For this initial study, we rely on 40 textile samples. We cut out pieces $40\times17$ cm in size. We have 10 polyester and wool samples, and 20 cotton samples. Polyester samples are woven and wool samples are knitted. Out of the 20 cotton samples, 10 are woven and 10 are knitted. We cut the pieces so that the yarn direction is along the axis of pulling, with the more elastic direction of stretching being 
orthogonal to the axis of pulling as can be seen in Fig. \ref{fig:streching_explanation}. 
Two Franka Emika Panda arms are equipped with wrist-mounted Optoforce 6-axis Force-Torque (FT) sensors and flat 3D-printed grippers.

For data collection, we aimed to capture the samples' properties by means of two exploratory procedures \cite{explor_proc}, \emph{pulling} and \emph{twisting}, and investigate if they are consistently classifiable. To further analyse different data collection strategies, we decided to let just one arm move (also called \textit{active arm}) while the other one is kept still (\textit{passive arm}), see Fig. \ref{fig:streching_explanation}. A precise definition of these two manipulation actions is the following:
 \begin{itemize}
\item {\textbf{Pull}} The active arm exerts force on the sample by steadily moving 2 cm away from the static passive arm, maintaining a motion direction parallel to the grasping plane.
 
 \item {\textbf{Twist}} The active arm's end-effector rotates 90 degrees, while the arm is pulling to ensure the sample is stretched adequately to capture its reaction to torsion.
  \end{itemize}
 
Each textile sample was held with a grasping force of 20N by both robot arms and pulled and twisted 20 times. As each sequence of pulling and twisting may result in a slight offset of the contact point, we  re-positioned  the sample in the hand to the original starting points after every five pulls/twists.
Force and torque signals were recorded for a duration of $2$s from each sensor at a frequency of $1$kHz.  Thus, for each textile sample, we have 20 examples, one for each arm, with $2\times1000\times6$ raw FT measurements.

\subsection{Dataset design}

We first sub-sampled the raw measurements for each FT dimension. We performed average downsampling to $150$ values as a trade-off between noise reduction and information loss.
Given these, we build 6 datasets:
\begin{itemize}
    \item $\mathcal{D}_{active}^{twist}$, $\mathcal{D}_{active}^{pull}$, $\mathcal{D}_{passive}^{twist}$, $\mathcal{D}_{passive}^{pull}$: 4 datasets corresponding to the two actions for each arm individually.
    \item $\mathcal{D}^{twist}$ and $\mathcal{D}^{pull}$: 2 datasets
    corresponding to the two actions and the integrated measurements from the two arms.
    \end{itemize}
    
In summary, the datasets was labelled to represent the samples of the taxonomy in Fig. \ref{fig:taxo} highlighted in red.
Therefore, by considering the "Production Method" branch we obtained 2 classes: woven and knitted, while on the "Yarn/Fiber Material" branch, we obtained 3 classes: wool, polyester and cotton.

\section{Data Visualization and Dataset Insights}

We first inspect the generated datasets to assess to what extent the collected data, labelled according to the proposed taxonomy, are representative for classification. To this end, we employ t-SNE\cite{maaten2008visualizing} and project the datasets into a two-dimensional space. More specifically, we want to answer the following questions:
\begin{itemize}
    \item Can the generated data and employed actions show a clear distinction between woven and knitted textiles?
    \item Is there a difference between pulling and twisting, in terms of how informative they are, for the woven vs knitted classification?
    \item What is the effect of fiber type on the classification performance and can we distinguish not only the production method, but also the fiber type given our datasets? 
\end{itemize}

\begin{figure}[!htb]
  \centering
\includegraphics[width=\linewidth]{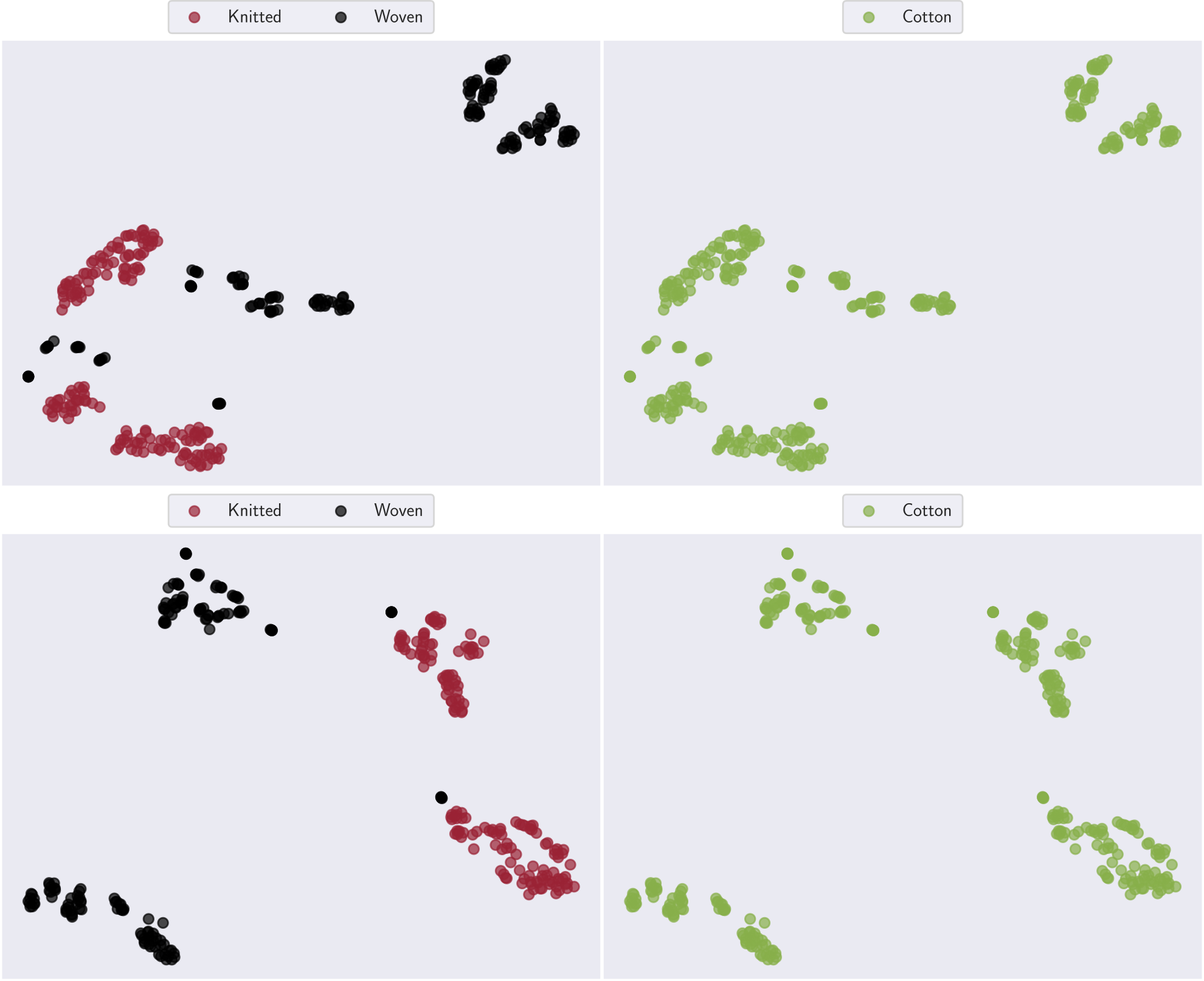}
  \caption{Effect of splitting construction method on Cotton samples. t-SNE plot of measurements obtained from the active (top row) and passive (bottom row) arms during pulling Cotton samples.  }\label{fig:t_snes1}
\end{figure}

\subsection{Insight 1: Production Method}
\label{ssec:ins1}
First, we visually inspect if the datasets are representative of the production method. Fig.\ref{fig:t_snes1} shows the distribution of the data projected in 2D when only cotton is considered, where the left side of the figures shows how the data is separated by the production method, woven (also called Cotton-Twill) and knitted (also called Cotton-Jersey). The top row corresponds to the active arm and the bottom row to the passive arm during a pulling trial. From the figure we observe that by considering the actual production method we obtain clearly structured groups of samples that would have been otherwise masked by categorizing them as the same material.

\subsection{Insight 2: Pulling vs Twisting}
\begin{figure}[!htb]
  \centering
  \includegraphics[width=\linewidth]{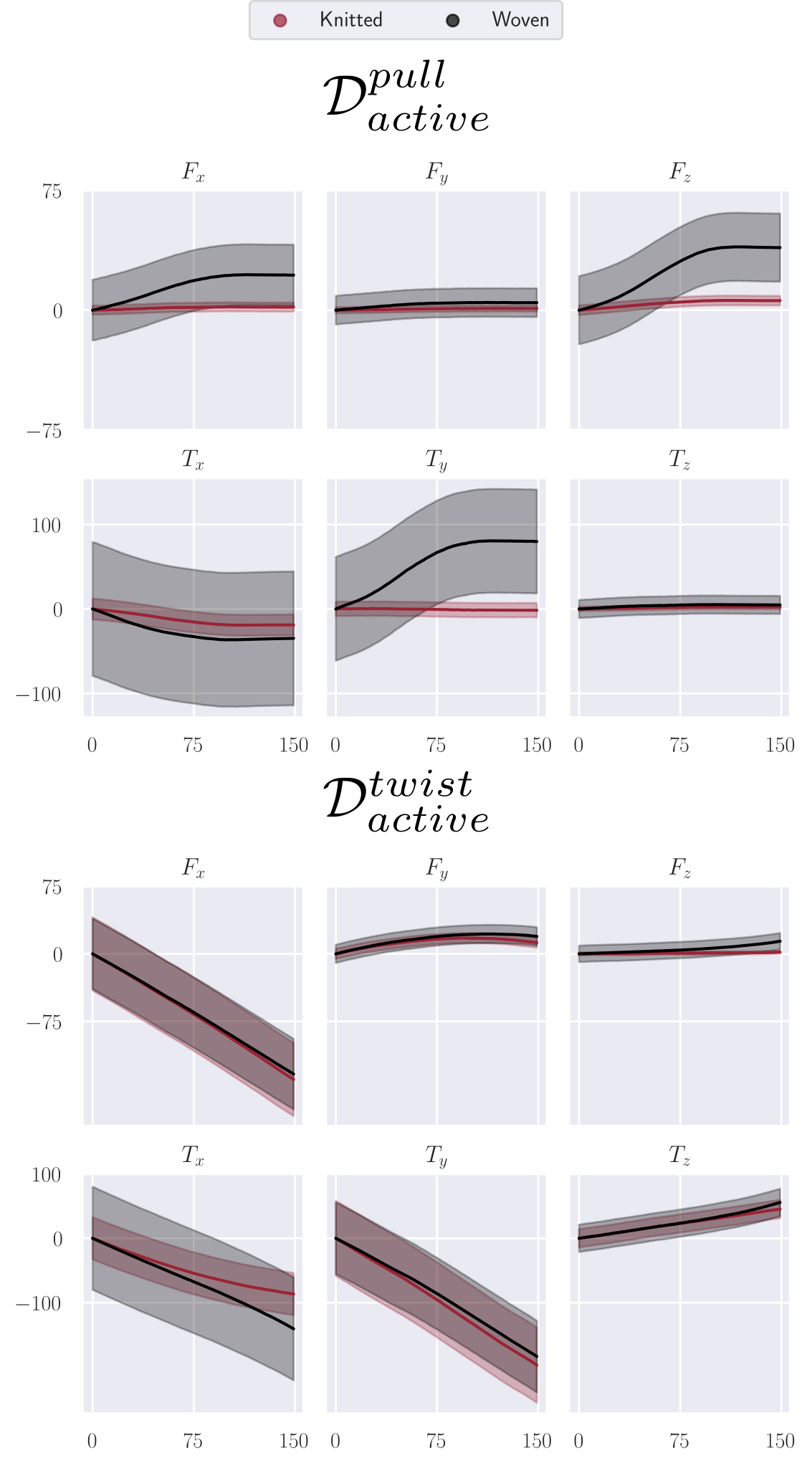}
  \caption{{Pull vs twist measurements on the active arm.}
   }\label{fig:pull_vs_twist}
\end{figure}

Second, we assess whether there is an advantage in using both pulling and twisting. As a first step, Fig.~\ref{fig:pull_vs_twist} shows that measurements for the two actions exhibit different behavior. For example, the force measurements during pulling exhibit more variety than they do for twisting, indicating that they may carry more information for the different classes.

To further examine how important the different measurements are and how they can affect classification, we 
train a simple SVM~\cite{cortes1995support} classifier to predict the sample's class, while trained on individual signals ($F_x$, $F_y$, $F_z$, $T_x$, $T_y$, $T_z$).
The classifier has a linear kernel and it is implemented with Scikit-learn~\cite{scikit-learn}. Table~\ref{tab:res_manip} shows the test set accuracy based on the individual signals for the \emph{pull} (left) and the \emph{twist} (right) strategies of the \emph{active} arm, when learning on a material-based dataset and a construction-based ones.

\begin{table}[h!]
\centering
{\renewcommand{\arraystretch}{1.5}%
\begin{tabular}{|l|c|l|l|l|}
\hline
\multicolumn{1}{|c|}{\multirow{2}{*}{\textbf{Signal}}} & \multicolumn{2}{c|}{\emph{\textbf{Pull}}}         & \multicolumn{2}{c|}{\emph{\textbf{Twist}}} \\ \cline{2-5} 
\multicolumn{1}{|c|}{}                        & \multicolumn{1}{l|}{Material} & Construction & Material        & Construction        \\ \hline
     $F_x$     &  $30$\%              & $82$\%     &      $45$\%        &     $57$\%        \\ \hline
     $F_y$     & $32$\%                &   $63$\%   &       $38$\%       &      $52$\%       \\ \hline
    $F_z$      & $70$\%           &  $100$\%    &      $53$\%        &      $73$\%       \\ \hline
    $T_x$    &   $42$\%         &  $75$\%    &      $38$\%        &      $63$\%       \\ \hline
    $T_y$     &   $44$\%       &   $83$\%   &      $38$\%        &      $63$\%       \\ \hline
    $T_z$        &  $40$\%     &  $53$\%    &      $45$\%        &      $59$\%       \\ \hline
    All   &   $80$\%     &  $100$\%    &      $70$\%        &      $87$\%       \\ \hline
\end{tabular}
}
\caption{SVM test set performance based on the individual signals for a dataset with 3 classes for material distinction and 2 classes for the production methods. }
\label{tab:res_manip}
\end{table}

From the SVM results, we observe that for the material-based classification, the accuracy scores for the two actions are comparable and rather low. However, when considering the proposed labels, the accuracy increases and in almost all signal cases, pulling outperforms twisting.

These results reinforce that following construction-based taxonomy is advantageous as well as the notion that the sensory feedback varies a lot depending on how textile is manipulated. It is therefore of fundamental importance to understand how to choose the proper exploration strategy. Moreover, as mentioned in Section \ref{sec:taxonomy}, the way in which textile threads are interlocked leads to different elastic properties. Knitted textiles, for example, can be stretched lengthwise or along the horizontal direction. Woven textiles instead, are usually not stretchable, apart from a bias direction that for example, for denim is the diagonal one. All these concepts play an important role in classification and highly increase the complexity of the task.


\subsection{Insight 3: Fiber Material vs Production Method}
\begin{figure}[!ht]
  \centering
  \includegraphics[width=\linewidth]{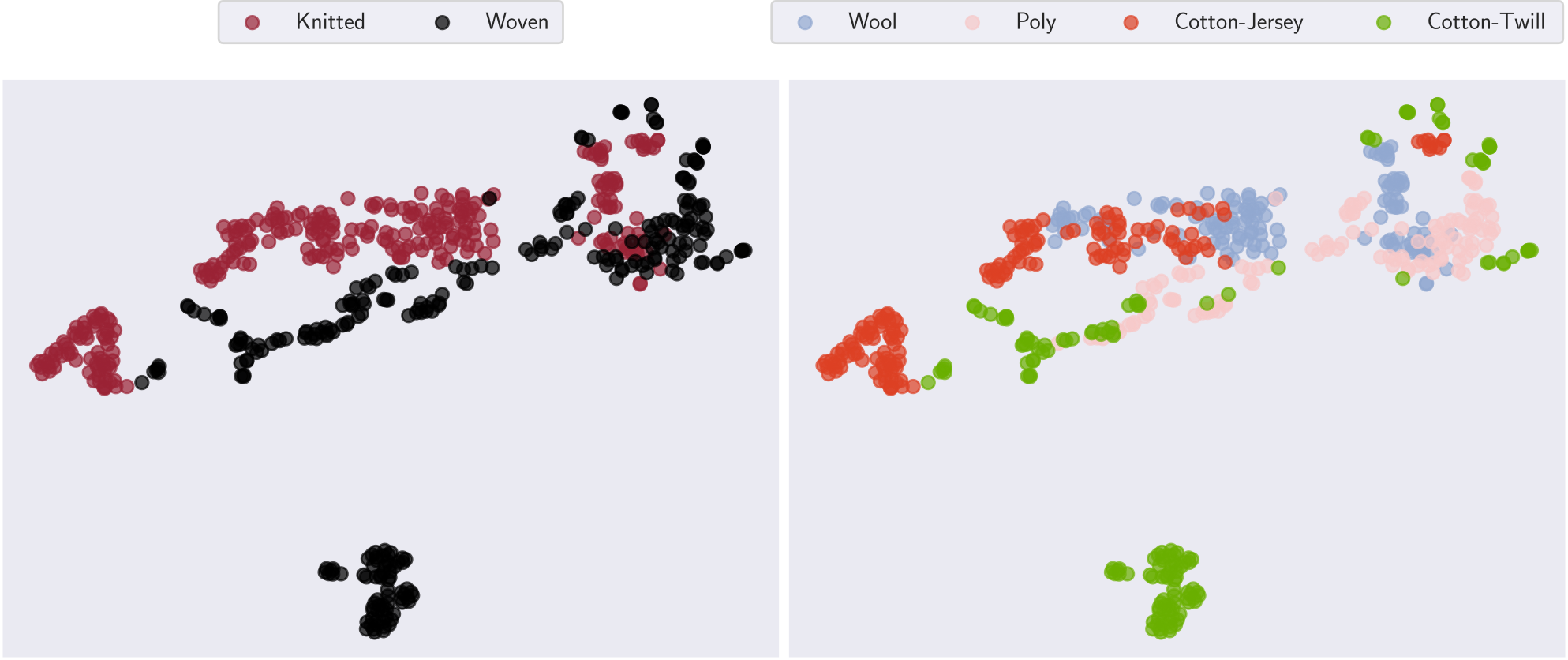}
  \caption{{Twist on the passive arm: visualization with respect to left) production method, right) fiber material.}
   }\label{fig:t_snes6}
\end{figure}

Lastly, we assess to what extent the fiber type can be identified in addition to the production method. An example of this can be seen in Fig.~\ref{fig:t_snes6} for dataset $\mathcal{D}_{passive}^{twist}$ that depicts the difference between samples categorized using just their production method (knitted or woven) and when they are labelled based on the material with the further distinction of the Cotton class, which is split into Cotton-Twill (woven) and Cotton-Jersey (knitted) to reflect its production methods. We can see that Cotton-Twill visually belongs to a separate cluster as observed in section \ref{ssec:ins1}. It can be also noticed that some of the Cotton-Twill samples are closer to Polyester as to Wool,  while Cotton-Jersey is closer to Wool than to Polyester. 

\begin{figure}[!htb]
  \centering
  \includegraphics[width=\linewidth]{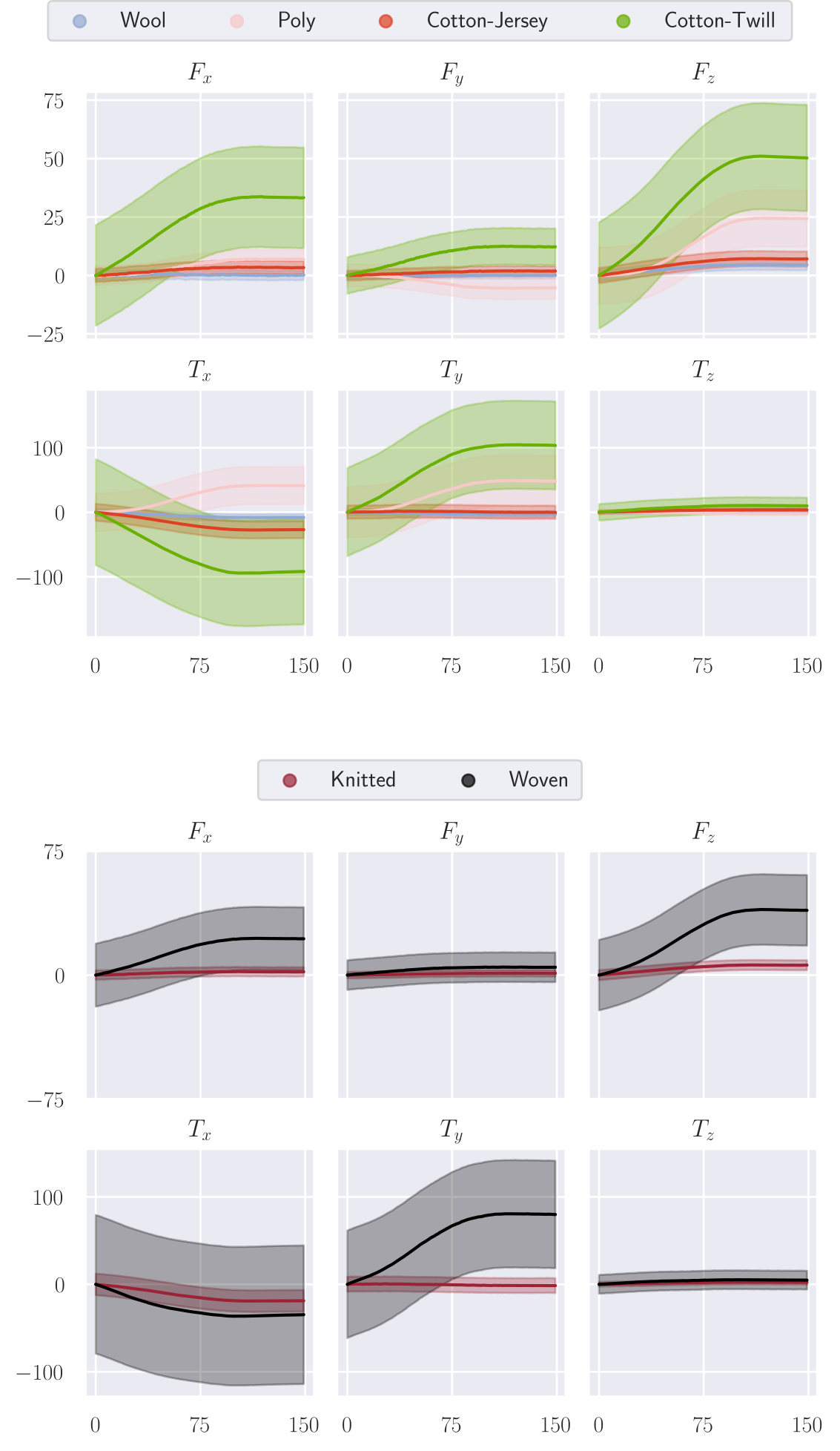}
  \caption{{Mean and standard deviation of the measurements sensed by the active arm while pulling. }
   }\label{fig:mean_std}
\end{figure}

We can also observe the effect of the proposed taxonomy on the individual signal level, by considering the dataset $\mathcal{D}_{active}^{pull}$ for example. 
Fig.~\ref{fig:mean_std} depicts how the split of Cotton by construction method highlights the difference between the mean force used for Cotton-Twill and Cotton-Jersey at the end of the pulling action, further showing the necessity of splitting the Cotton class. Moreover, besides the detectable distinction among Polyester, Wool, Cotton-Twill and Cotton-Jersey signals, woven materials keep being the ones with higher mean force while knitted ones are in general less tension-resistant, reflecting the behaviour of the construction methods.

\section{Classification and Interpretability}
The next step is to assess the classification performance using a more complex architecture, like a CNN model. The input of the network are vectors of the six concatenated FT measurements, the network consists of four 2D convolutional layers and its activations are rectified linear units (ReLU). Furthermore, we adopt rectangular kernels of size $5\times 1$ which convolve across measurements of the same signal as done in \cite{deepLearningVisualHaptic}. The output sizes of the convolutional layers are respectively $24, 12, 8$ and $4$.

 The features learned from the last convolutional layer are flattened and fed to a fully connected layer with 48 hidden neurons. The outputs of this block are the predicted class probabilities. We also consider the case of the joined measurements for the \emph{active} and \emph{passive} arm, using the same architecture but adjusting the size of the fully-connected layer to accommodate the $150\times 12$ signal.

\begin{table}[h!]
\centering
{\renewcommand{\arraystretch}{2}%
\begin{tabular}{|l|c|c|}
\hline
\multicolumn{1}{|c|}{\textbf{Input Dataset}} & \textbf{Materials} & \textbf{Construction} \\ \hline
 $\mathcal{D}_{active}^{pull}$ & $87.5 \%$ & $100\% $ \\ \hline
$\mathcal{D}_{passive}^{pull}$ & $85.8\%$ & $96.7 \%$ \\ \hline
$\mathcal{D}_{active}^{twist}$ & $76.7 \%$ & $95.0\%$ \\ \hline
$\mathcal{D}_{passive}^{twist}$ & $78.3 \%$ & $89.2 \%$ \\ \hline
$\mathcal{D}^{pull}$ &  $95.0\%$ & $100\%$  \\ \hline
$\mathcal{D}^{twist}$ &  $79.0\%$ & $91.7\%$  \\ \hline

\end{tabular}
}
\caption{Test accuracy with a CNN model for all the different datasets when following the material-based labelling and the proposed, construction-based one.}
\label{tab:class_res}
\end{table}

We partitioned each dataset into 90/10 train/test splits. Table~\ref{tab:class_res} summarizes the classification results. Firstly, we observe that the construction-based labeling outperforms the material-based one for any dataset or action. More specifically, using our taxonomy both actions provide enough information for accurate predictions. However, twisting  is consistently less accurate than pulling for all labeling and datasets considered. 

These observations are validated in the case of the joined datasets with $\mathcal{D}^{pull}$ achieving excellent performance in distinguishing the construction method of the textile, leading to the conclusion that pulling is a better option for classification.
 
\subsection{Interpretability and measurement assessment} 


To further examine the effect of the different measurements for classification, we interpret the results from the CNN model through GradCAM \cite{GradCam}. GradCAM is an interpretability technique that produces visual explanations in the form of heatmaps that portray which parts of the input contribute the most to the predicted label. We follow the same methodology as in \cite{shamelessMitsioni} to produce and inspect the contribution of each feature in samples from datasets $\mathcal{D}^{pull}$ and $\mathcal{D}^{pull}_{active}$ of Table \ref{tab:class_res}.

\begin{figure}[!htb]
  \centering
  \includegraphics[width=0.9\linewidth]{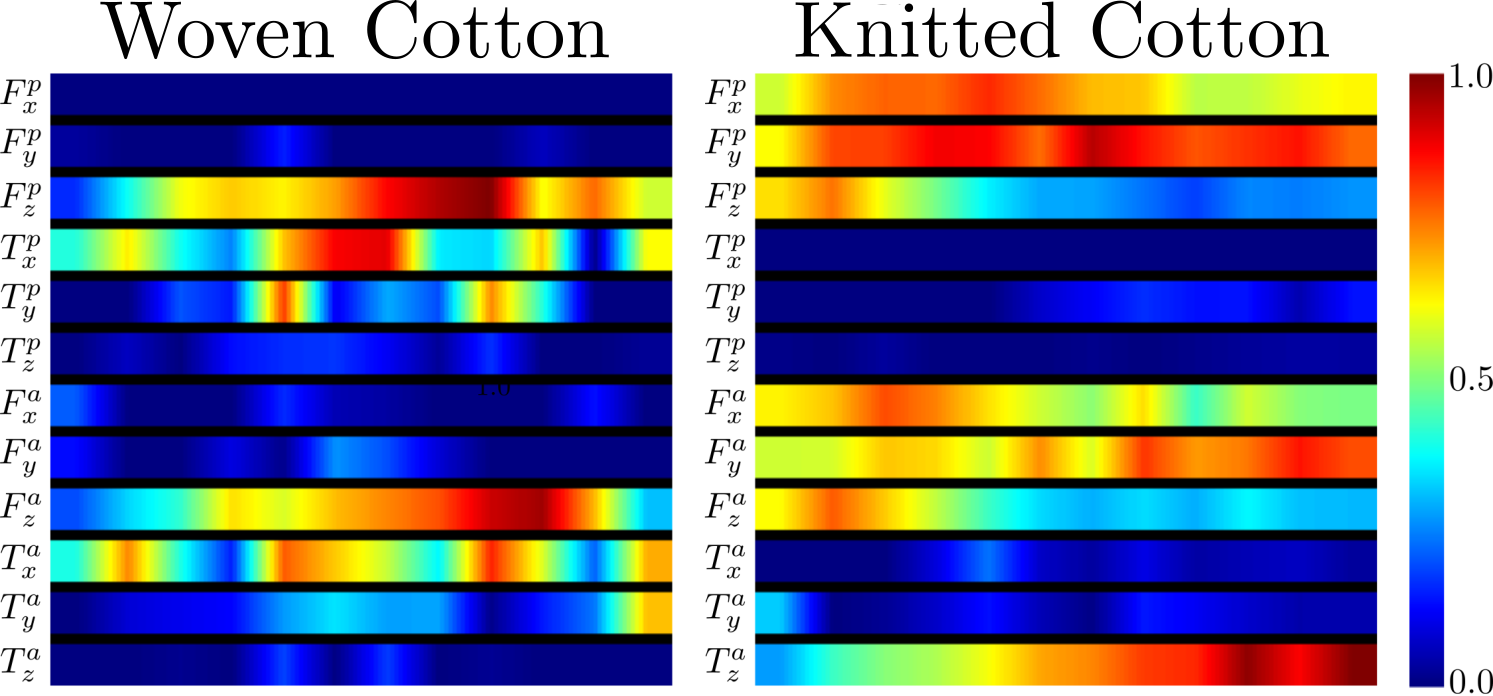}
  \caption{Heatmaps of feature importance for the classification using dataset $\mathcal{D}^{pull}$. The intensity of each row (with red being the most important) denotes what the network focuses on to classify the sample. 
   }\label{fig:heat_12}
\end{figure}

An example of the heatmaps can be seen in Fig. \ref{fig:heat_12} for two correctly classified samples of woven cotton and knitted cotton from the dataset $\mathcal{D}^{pull}$. Every row corresponds to a different measurement channel and its color is defined by how important it is for the prediction. The importance is scaled between 0 and 1 and follows the colormap on the right of the images. 
Fig. \ref{fig:heat_12} shows that for both cotton instances, the network is focusing on the same features between the passive and the active arm. Concretely, for woven cotton, the forces on both axes $Z$ are the most important features, followed by the torques $T^p_x, T^a_x$ and some parts of $T^p_y, T^a_y$.  However, for knitted cotton, the network utilized all the force measurements for both arms and the torque on axis $Z$ for the active arm. These results indicate that even when utilizing the material based-labeling, the CNN network focuses on different patterns when classifying samples of the same material but different construction methods.

Finally, we inspect two classification results from dataset $\mathcal{D}^{pull}_{active}$ when it is labelled according to the proposed taxonomy. The left heatmap corresponds to a correct classification of a woven sample and the right heatmap on the correct classification of a knitted one. 
The important features agree with the intuition gained from Table \ref{tab:res_manip} as the decisions are heavily based on the ones highlighted 
also by the SVM, namely forces $F_z$ and torques $T_x, T_y$.

 \begin{figure}[!htb]
  \centering
  \includegraphics[width=0.9\linewidth]{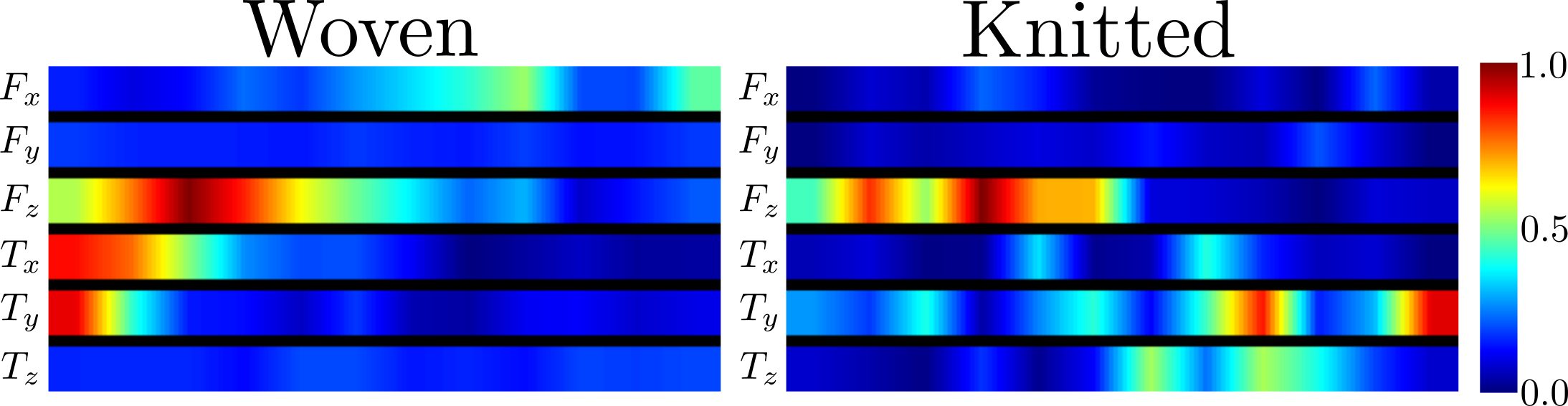}
  \caption{Heatmaps for the dataset $\mathcal{D}^{pull}_{active}$: activation for left) a correctly classified Woven sample, right)  correctly classified Knitted sample.
   }\label{fig:heat_6}
\end{figure}

We note that the visualization shows only where the neurons of the network are most active for single examples. It is possible for a network to construct multiple patterns to classify the same class, making generalisation difficult. We can however, observe that certain measurements are more important for classification than others, which is a valuable insight when designing future active exploration strategies.

\section{Discussion and Conclusion}
In this work, we outlined a textile taxonomy and showed our initial results on textile sample classification using pulling and twisting actions. The focus of the study was to assess to what extent a taxonomy used in textile industry is a viable model to structure robotic interaction and provide a basis for a whole new area of structured studies of this class of deformable materials. 

We provided insights into how a combination of different actions and FT measurements vary with respect to textile production method and fiber material. Pulling and twisting, as inspired by the human interaction with textile, are viable choices of actions and these provide relevant information for classification. One interesting question that arises is what other actions can be potentially employed and to what extent dexterous hand/finger motion could be exploited in addition to pulling and twisting. Some previous work demonstrated the use of specialized fingers and sensors for this purpose and it is yet to be seen to what extent we can consider such solutions to become commercial. 

Combining multiple actions, as well as passive and active interaction, is also an interesting aspect to be explored. We may start with pulling/stretching and based on the first step classification, subsequent routines may be performed more suitably for identifying categories of interest, such as for example fiber material, elasticity, whether the textile is wet or dry, etc. Here, reinforcement learning may be used to learn actions that maximize the utility of the sensor readings for discriminating various textile properties. 

We also performed an initial study using visual feedback under pulling and twisting. However, for the considered categories, regular cameras do not provide enough resolution to bring sufficient information on the production method or the fibre type. One could potentially rely on the reflectance properties of textile materials, but most of the works in this area that stem from the computer vision community, are not applicable in uncontrolled settings that would occur in real-life applications. One idea that arose when conducting the study was the fact that creases and wrinkles on the textile fabric may be a useful feature to exploit for certain applications. When pulling or stretching the fabric in many different directions, creases and wrinkles will vary dependent on the properties of the textile: dense and hard textile creases differently from soft and thin textile. In such cases, integrating vision and FT may be useful. Careful consideration  on what visual features are used needs to be taken into account. For example, using flow-based methods~\cite{scholz2004cloth} or specified wrinkle detectors~\cite{takamatsu2019study} dealing with various texture properties may be considered. 

An additional important aspect to be considered is the ability to assess how textile properties change over time. Certain textiles are made to be more durable, fibers are blended, their use and handling in terms of washing, ironing, folding, will affect how clothing items deteriorate over time. In other words, the information of the fiber content usually available on the label sewn on the clothing item,  may be helpful but it is not fully relevant. For example, a T-shirt made out of cotton, may be more elastic and thicker when new, and rather thin and almost non-elastic after many washings. Thus, its handling in terms of washing and ironing will be different, as well as one may decide to keep or reuse a newer one, and recycle a well-used one.  

The outlined taxonomy, visualization, CNN classification and measurement interpretability are important tools that can provide more insight into the difficulty of the considered problem. The taxonomy provides a structured approach to study textile materials and has not been previously considered in the area of robotics. We also need an approach that brings the robotics community closer to textile production industry and this is one way of achieving that. We provided several examples of how 
the generated textile material classes are a viable approach and how these can be studied together with actions such as pulling and twisting. 

Initial classification results using deep neural networks show a good potential and we will build on these with a more extensive database of samples, actions and multimodal sensory feedback. 
More specifically, we will study a richer set of pulling actions, with samples of different sizes also considering standardized textile for the purpose of repeatability, reproducibility and replicability. We believe that this study is an important step toward a more robust and versatile textile handling and manipulation for applications such as various household tasks, assisted dressing and recycling.

\bibliography{references}
\bibliographystyle{IEEEtran}

\end{document}